%% file: main.tex
\documentclass[10pt,twocolumn,letterpaper]{article}

\usepackage[pagenumbers]{cvpr} 

\usepackage{graphicx}
\usepackage{amsmath}
\usepackage{amssymb}
\usepackage{booktabs}
\usepackage[table,xcdraw]{xcolor}
\DeclareMathOperator*{\argmaxA}{arg\,max}

\usepackage[pagebackref,breaklinks,colorlinks]{hyperref}
\usepackage{acronym}
\usepackage{float}
\usepackage{subcaption}
\usepackage{paralist}

\usepackage[capitalize]{cleveref}
\crefname{section}{Sec.}{Secs.}
\Crefname{section}{Section}{Sections}
\Crefname{table}{Table}{Tables}
\crefname{table}{Tab.}{Tabs.}

\def\cvprPaperID{2134} 
\def\confName{CVPR}
\def\confYear{2023}

\acrodef{cv}[CV]{Computer Vision}
\acrodef{nlp}[NLP]{Natural Language Processing}
\acrodef{llm}[LLM]{Large Scale Language Model}
\acrodef{vlp}[VLPs]{Vision-Language Pre-trained Models}
\acrodef{vl}[V\&L]{Vision-Language}
\acrodef{mlm}[MLM]{Masked Language Modeling}
\acrodef{itm}[ITM]{Image-Text Matching}
\acrodef{ikm}[IKM]{Image-Knowledge Matching}
\acrodef{iec}[IEC]{Image Edit Checking}

\begin{document}

\title{GIVL: Improving Geographical Inclusivity of \\ Vision-Language Models with Pre-Training Methods}

\author{Da Yin$^{1}$ \quad Feng Gao$^{2}$ \quad Govind Thattai$^{2}$ \quad Michael Johnston$^{2}$ \quad Kai-Wei Chang$^{1,2}$ \\
$^1$ University of California, Los Angeles \qquad $^2$ Amazon Alexa AI\\
{\tt\small \{da.yin,kwchang\}@cs.ucla.edu, \{fenggo,thattg,mjohnstn\}@amazon.com}
} 
\twocolumn[{%
\renewcommand\twocolumn[1][]{#1}%

\maketitle
\vspace{-15pt}
\begin{center}
    \centering
    \captionsetup{type=figure}
    \includegraphics[width=0.8\textwidth, trim=0 0 0 0, clip]{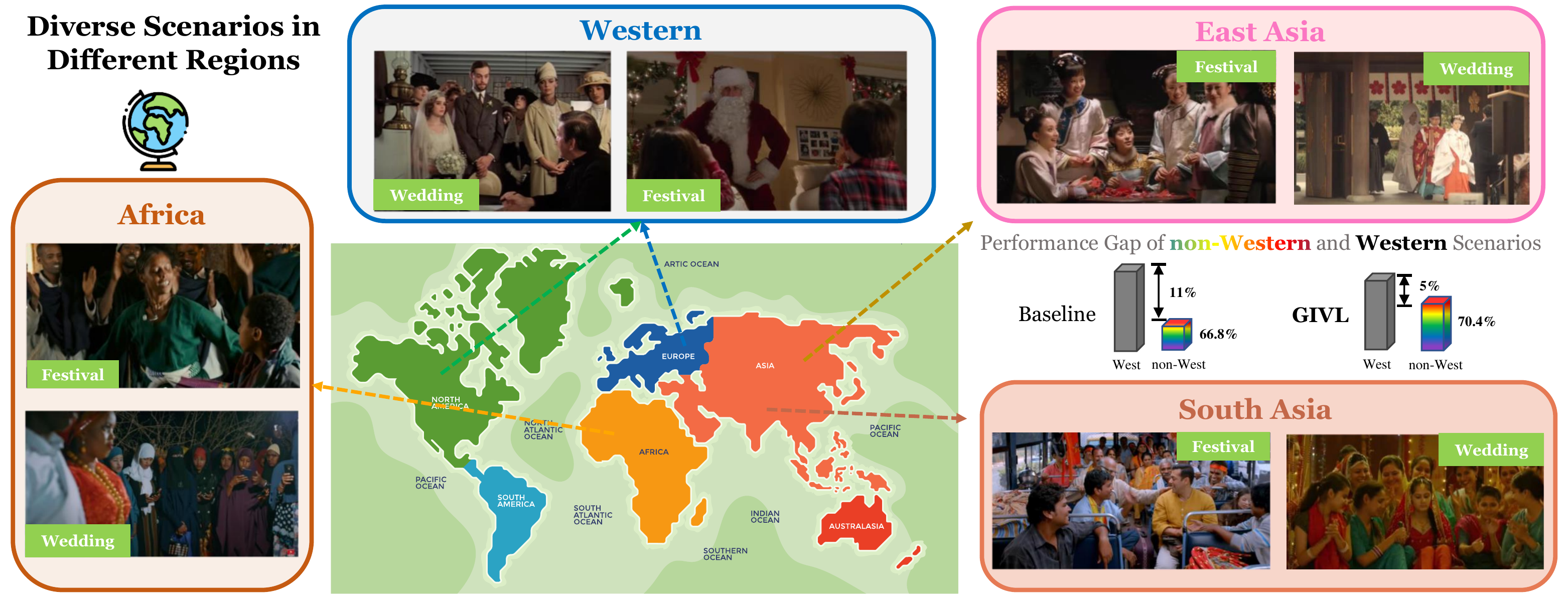}
    \vspace{-3pt}
    \captionof{figure}{Scenarios around the world including festivals and weddings. Even the same scenarios have distinct visual characteristics across regions (a.k.a. geographically diverse). Compared with prior \ac{vlp}, GIVL achieves much better performance on non-Western data in GD-VCR\cite{yin-etal-2021-broaden}. GIVL can also make the gap between Western and non-Western cases much closer.}
\label{intro-fig-1}
\end{center}
}]

\vspace{-3pt}
\begin{abstract}
\vspace{-15pt}
A key goal for the advancement of AI is to develop technologies that serve the needs not just of one group but of all communities regardless of their geographical region.
In fact, a significant proportion of knowledge is locally shared by people from certain regions but may not apply equally in other regions because of cultural differences. 
If a model is unaware of regional characteristics, it may lead to performance disparity across regions and result in bias against underrepresented groups. We propose GIVL, a \textbf{G}eographically \textbf{I}nclusive \textbf{V}ision-and-\textbf{L}anguage Pre-trained model. There are two attributes of geo-diverse visual concepts which can help to learn geo-diverse knowledge: 1) concepts under similar categories have unique knowledge and visual characteristics, 2) concepts with similar visual features may fall in completely different categories. Motivated by the attributes, we design new pre-training objectives Image-Knowledge Matching (IKM) and Image Edit Checking (IEC) to pre-train GIVL. Compared with similar-size models pre-trained with similar scale of data, GIVL achieves state-of-the-art (SOTA) and more balanced performance on geo-diverse V\&L tasks.
\end{abstract}

\input{sections/intro}
\input{sections/related}
\input{sections/method}
\input{sections/experiments}
\input{sections/conclusion.tex}

{\small
\bibliographystyle{ieee_fullname}
\bibliography{references}
}

\clearpage

\input{sections/appendix.tex}

\end{document}

%% file: sections/intro.tex
\section{Introduction} \label{sec:intro}

\acf{vlp} \cite{li2019visualbert,lu2019vilbert,zhang2021vinvl,li2021align,dou2022empirical} have achieved remarkable performance on \acf{vl} tasks including visual question answering\cite{balanced_vqa_v2,hudson2018gqa,gao2022transform}, image-text retrieval\cite{li2020unicoder}, and image captioning\cite{lin2014microsoft,karpathy2015deep}. Pre-trained with large-scale corpora of image-text pairs, \eg COCO\cite{lin2014microsoft}, OpenImages\cite{kuznetsova2020open}. \ac{vlp} are capable of learning multi-modal representations and can be effectively fine-tuned on downstream \ac{vl} tasks. 

While \ac{vlp} can solve a broad range of \ac{vl} tasks, to deploy \ac{vlp} in real-world applications, it is essential to consider the geographical inclusivity\footnote{We use regions as a proxy to estimate inclusivity of V\&L models. People in the same regions may have different cultures and traditions.} of \ac{vlp}. Because of geographic differences, images from different regions embody a large amount of knowledge that is locally shared but cannot be applied in other regions, \ie geographically diverse. For example, in Figure~\ref{intro-fig-1}, the festivals in different regions look different. 

Ideally, a geographically inclusive VLP should be capable of achieving comparable performance over all the images, regardless of their origins. 
However, current \ac{vlp} does not perform equally well on data from different regions. For example, prior works\cite{liu-etal-2021-visually,yin-etal-2021-broaden} show that on geo-diverse \ac{vl} tasks, there is nearly a 20\% performance discrepancy between Western and East Asian images when current \ac{vlp} are applied. To combat such geographical bias, we aim to design methods to make \ac{vlp} achieve more balanced performance across regions.

One solution to mitigating bias is to obtain diverse task-specific annotations for each region and fine-tune \ac{vlp} on the new annotations. However, according to \cite{ipeirotis2010demographics}, most Amazon MTurk annotators are from US and India, and may be unfamiliar with the cultures of other regions. Thus, it is unrealistic to obtain large-scale geo-diverse annotations even in such a popular crowdsourcing platform. 

Pre-training a \textit{unified} VLP with large-scale \textit{unannotated} geo-diverse images and corresponding knowledge could make the VLP a foundation to provide more generalizable representations and help to transfer on comprehending images from various regions easier. 
In this paper, we propose \textbf{GIVL}, a \textbf{G}eographically \textbf{I}nclusive \textbf{V}ision-and-\textbf{L}anguage Pre-trained model. We focus on \textbf{how to encourage GIVL to better learn geo-diverse knowledge on images from different regions during its pre-training stage}.

We observe two attributes of geo-diverse visual concepts that can contribute to learning geo-diverse knowledge: 

\vspace{3pt}
\noindent \textbf{A1: Concepts under similar categories have unique knowledge and visual characteristics.} For example, traditional Western and Chinese festivals, like \textit{Christmas} and \textit{Chinese New Year} in Figure~\ref{intro-fig-1}, are held with different rituals and their decoration style differs as well. It is necessary for GIVL to learn the difference between their corresponding knowledge and precisely distinguish these visual concepts. On the other hand, \textit{Christmas} and \textit{Chinese New Year} are both festivals. Learning the commonalities of visual concepts (e.g., both images in Figure~\ref{intro-fig-1} belong to the same category ``festival'') would help model connect Western and non-Western concepts and contribute to more effective transfer on geo-diverse images.

\begin{figure}[t]
\centering
\includegraphics[width=0.6\columnwidth]{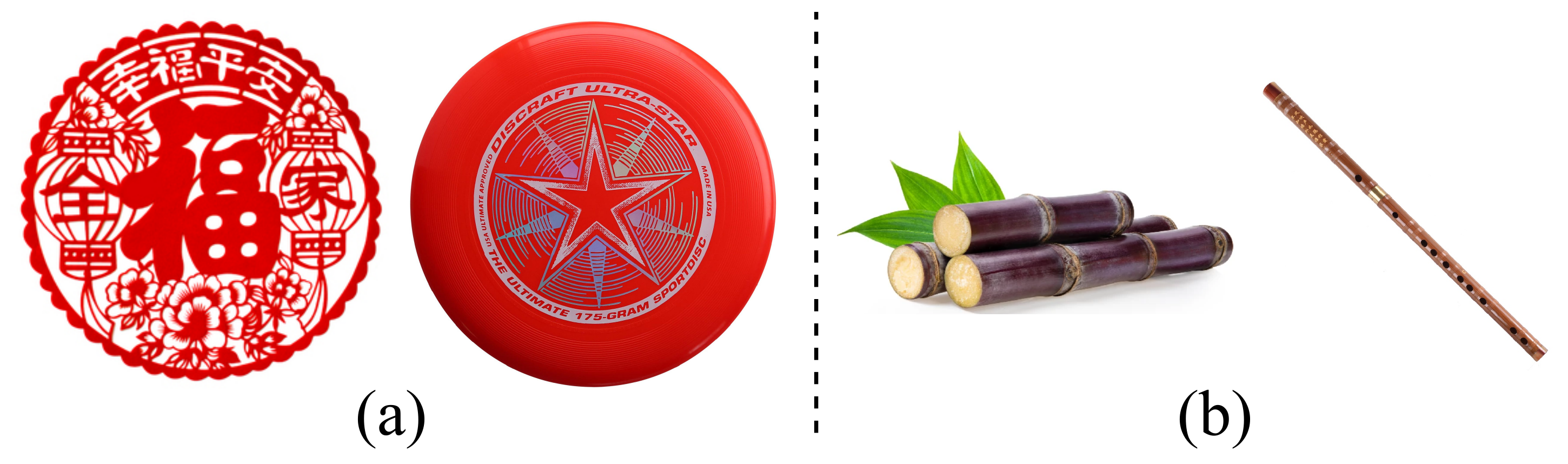}
\caption{Example of Chinese paper cuttings and red frisbee (left) and sugar cane and flute (right). Different visual concepts may share similar visual characteristics, but they may have completely different functionalities.}
\label{intro-fig-2}
\vspace{-4pt}
\end{figure}

\vspace{3pt}
\noindent \textbf{A2: Concepts with similar visual features may lie in completely different categories.} In Figure~\ref{intro-fig-2}, \textit{Chinese paper cuttings} share visual features (e.g., color, shape) with \textit{red frisbee}. Similarly. \textit{sugar cane} and \textit{flute} share visual features. However, these concepts are not related to each other. Since geo-diverse images cover a broader range of visual concepts, differentiating visually similar concepts given visual contexts is also essential.

To this end, besides common objectives \acf{mlm} and \acf{itm} for pre-training VLPs, we propose two additional pre-training objectives, \textbf{\acf{ikm}} and \textbf{\acf{iec}}. \ac{ikm} is used to learn the alignment between images and corresponding textual knowledge in Wikipedia. It requires GIVL to not only judge if the input textual knowledge matches input images, but also identify whether the visual concepts described in input knowledge falls into similar categories of the concepts in input images. This encourages GIVL to learn corresponding relationship between knowledge and images as well as recognize similarity among geo-diverse visual concepts. \ac{iec} is proposed to identify whether a visual concept in input image is replaced by another concept that is visually similar but lies in an irrelevant category (see Fig.\ref{fig:model} for an example). It enables GIVL to capture nuances between visually similar concepts after the replacement given visual contexts. 


Our contributions and empirical results are as follows:
 \begin{itemize}
    \item By considering the attributes of geo-diverse visual concepts, we propose two novel \ac{vl} pre-training objectives Image-Knowledge Matching (IKM) and Image Edit Checking (IEC) that can greatly improve the geographical inclusivity of \ac{vlp}. 
    \item Compared with similar-size VLPs pre-trained with similar scale of data, GIVL\footnote{Code and model checkpoint will be released.} achieves state-of-the-art (SOTA) and more balanced performance over different regions on geo-diverse \ac{vl} tasks including MaRVL~\cite{liu-etal-2021-visually}, GD-VCR~\cite{yin-etal-2021-broaden} and WIT Image-Text Retrieval~\cite{srinivasan2021wit}. For geo-diverse zero-shot image classification on Dollar Street dataset\footnote{Dollar street dataset is available at \url{https://github.com/greentfrapp/dollar-street-images}.}, GIVL outperforms VinVL~\cite{zhang2021vinvl} 26\%.
\end{itemize}


%% file: sections/related.tex
\section{Related Work} \label{related}

\paragraph{Vision-Language Pre-Trained Models (VLPs).}
\ac{vlp} \cite{lu2019vilbert,li2019visualbert,tan2019lxmert,li2020unicoder,li2020oscar,chen2020uniter,zhang2021vinvl,li2021align,shen2021much,dou2022empirical} are proposed to tackle tasks that require understanding of both images and texts. Following the paradigm of pre-training language models \cite{peters-etal-2018-deep,devlin-etal-2019-bert,radford2019language}, in common practice, \ac{vlp} use Transformer~\cite{vaswani2017attention} as the backbone and pre-train them with large-scale image-caption pairs. The commonly used image-text parallel data are from multiple sources including COCO~\cite{lin2014microsoft}, Flickr30K~\cite{young-etal-2014-image}, Conceptual Captions~\cite{sharma-etal-2018-conceptual} and OpenImages~\cite{kuznetsova2020open} datasets. Currently, VLPs have achieved remarkable performance on various \ac{vl} tasks including visual question answering\cite{balanced_vqa_v2,hudson2018gqa}, visual reasoning~\cite{suhr-etal-2019-corpus}, image captioning~\cite{lin2014microsoft}, and image-text retrieval \cite{lin2014microsoft,young-etal-2014-image}. Most recent works focus on scaling up VLPs; this paper studies an orthogonal but important concerns - how to leverage diverse knowledge to improve inclusivity of VLPs. 

\vspace{-9pt}
\paragraph{Geographical Bias.}
Geographical bias~\cite{shankar2017no,de2019does,wang2022revise,yin2022geomlama} is a severe problem in that AI applications have. 
Previous works~\cite{yin-etal-2021-broaden,liu-etal-2021-visually} reveal the fact that on geo-diverse V\&L tasks, the performance gap between non-Western and Western images is significant when using VLPs. Similarly, object recognition models' performance greatly drops on non-Western images~\cite{shankar2017no,de2019does}. Researchers~\cite{shankar2017no,de2019does,wang2022revise} find that one factor of the geographical bias is introduced by an imbalanced distribution of training data with respect to geographical location. They\cite{de2019does} observe that COCO and OpenImages, two widely used pre-trained corpora for \ac{vlp}, are amero-centric and euro-centric. 
Another reason behind the performance drop is that \ac{vlp} can understand basic visual information in images from different regions, but are less able to leverage geo-diverse knowledge and reason~\cite{yin-etal-2021-broaden}. 

\vspace{-9pt}
\paragraph{Geo-Diverse V\&L Tasks.}
GD-VCR~\cite{yin-etal-2021-broaden} studies whether a model can understand commonsense on geo-diverse images. V\&L models are required to select the correct answer from four answer choices given textual questions involving geo-diverse commonsense and the corresponding images. MaRVL~\cite{liu-etal-2021-visually} is another \ac{vl} task that requires visual reasoning with cultural knowledge of images from non-Western regions. It is formulated as a binary classification problem in which the model needs to judge whether a sentence correctly describes two images from non-Western regions. WIT image-text retrieval~\cite{srinivasan2021wit,bugliarello-etal-2022-iglue} is a standard multimodal retrieval task on geo-diverse Wikipedia images.

%% file: sections/method.tex
\section{Methods} \label{sec:method}
In this section, we introduce the pre-training method of GIVL in detail. Section~\ref{sec:method:prelim} provides preliminary of GIVL pre-training method including the definition of visual concept and category. Section~\ref{sec:method:objective} describes the four pre-training objectives. Section~\ref{sec:category} and~\ref{sec:locate} illustrate the process of acquiring essential information used to construct input contents for objectives \acf{ikm} and \acf{iec}. Specifically, Section~\ref{sec:category} shows how to extract visual concept name from an image caption and its category information from Wikipedia. Section~\ref{sec:locate} shows how to locate visual concept to corresponding detected objects in input image.

\subsection{Preliminary}\label{sec:method:prelim}
\begin{figure*}[ht]
    \centering
    \includegraphics[width=0.81\linewidth, trim=0 0 0 0, clip]{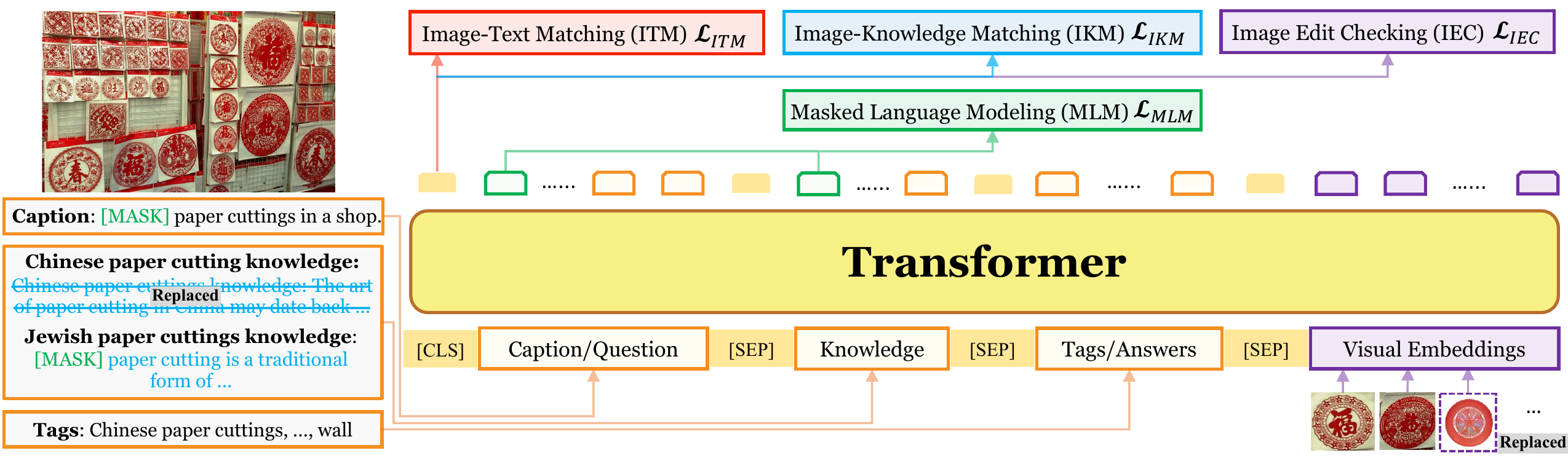}
    \caption{GIVL pre-training method with four pre-training objectives. The input image is about the visual concept \textit{Chinese paper cuttings}. The input knowledge is about \textit{Jewish paper cuttings} rather than \textit{Chinese paper cuttings}, but it is also the knowledge describing a visual concept that shares a similar category with \textit{Chinese paper cuttings}. Hence, for Image-Knowledge Matching (IKM) objective, the input contents belong to Type 3. Also, the visual concept \textit{Chinese paper cuttings} is replaced with a visually similar concept \textit{red frisbee}. Thus, for Image Edit Checking (IEC) objective, the input contents belong to Type 2.}
    \label{fig:model}
\end{figure*}

\paragraph{Definition of Visual Concept and Category.}
\texttt{Visual concept} is an object or scenario that an image mainly involves. For example, Figure \ref{fig:model} shows the visual concept of \textit{Chinese paper cuttings}.
Each specific visual concept corresponds to one general \texttt{category}. Each category covers various visual concepts having particular shared characteristics. For example, the category of visual concept \textit{Chinese paper cuttings} is \textit{art}. The \textit{art} category includes other visual concepts such as \textit{Jewish paper cuttings}. 
The extraction pipeline for visual concept and its category information will be introduced in Section~\ref{sec:category}.

\vspace{-9pt}
\paragraph{Pre-Training Corpus.}
To improve the geographical inclusivity of \ac{vlp}, we use Wikipedia Image-Text (WIT) dataset\cite{srinivasan2021wit} as a primary source of geo-diverse images. WIT contains 2.95M images in total\footnote{GIVL focuses on English-only \ac{vl} tasks. We only consider images with English captions, which only occupy 30\% out of the entire WIT.}. We also incorporate 0.22M commonly used \ac{vl} pre-training images from COCO\cite{lin2014microsoft}, Flickr30k\cite{young-etal-2014-image}, and GQA. Images in WIT dataset come with the corresponding Wikipedia sections that include the corresponding knowledge of WIT images. This knowledge\footnote{The knowledge of COCO, Flickr30K and GQA images is the first sentence in Wikipedia pages of the objects mentioned in captions.}, such as customs and history, is usually culturally related and not explicitly described in the images. Such knowledge plays a crucial role in helping \ac{vlp} understand visual concepts in geo-diverse images more comprehensively.

\vspace{-9pt}
\paragraph{Input for Pre-Training.}
We organize the input for GIVL pre-training as follows:
\begin{equation}    \mathrm{[CLS]}\;\mathbf{c}\;\mathrm{[SEP]}\;\mathbf{k}\;\mathrm{[SEP]}\;\mathbf{t}\;\mathrm{[SEP]}\;\mathbf{v},
\end{equation}
where $\mathbf{c}$ is either an image caption or a GQA question; $\mathbf{k}$ denotes the corresponding knowledge of the visual concept in input image $\mathbf{I}$; $\mathbf{t}$ is either tags of detected objects or a GQA answer; $\mathbf{v}$ is a list of visual embeddings generated from input image $\mathbf{I}$ by a ResNeXt-152 C4 detection model\cite{zhang2021vinvl}. $p_v$ is the name of the visual concept contained in image $\mathbf{I}$.

\subsection{Pre-Training Objectives for GIVL}\label{sec:method:objective}
We pre-train GIVL with four objectives: \acf{mlm}, \acf{itm}, \acf{ikm}, \acf{iec}. We introduce each pre-training objective as follows.

\vspace{-6pt}
\subsubsection{MLM and ITM Objectives}
\acf{mlm} is a learning objective prevalently used in \ac{vl} pre-training. Given the context of model inputs, GIVL needs to recover the tokens masked by $\mathrm{[MASK]}$. \ac{mlm} loss $\mathcal{L}_{MLM}$ is the average of all cross-entropy loss with respect to the probability of predicting the correct masked tokens given a vocabulary. 

\acf{itm} is another commonly applied objective that enables GIVL to learn the alignment between texts and images. Following VinVL\cite{zhang2021vinvl}, given an input image $\mathbf{I}$, we construct three types of input contents for $\mathbf{c}$ and $\mathbf{t}$. It is formulated as a 3-way classification task, $y^{c,t} \in \{0,1,2\}$, where 0 represents that $\mathbf{c}$ and $\mathbf{t}$ both match the input image $\mathbf{I}$; 1 means when $\mathbf{t}$ matches image $\mathbf{I}$ whereas $\mathbf{c}$ mismatches the image $\mathbf{I}$; 2 indicates $\mathbf{c}$ matches $\mathbf{I}$ but $\mathbf{t}$ mismatches $\mathbf{I}$. \ac{itm} loss is the cross-entropy loss with respect to the probability of predicting the type of input contents upon $\mathrm{[CLS]}$ representation.

\vspace{-6pt}
\subsubsection{Image-Knowledge Matching (IKM)}
\label{sec:ikm}
We propose \acf{ikm} to assist GIVL in learning knowledge of geo-diverse visual concepts. With the help of \ac{ikm}, we encourage GIVL to learn the corresponding knowledge of the visual concepts and discover connections between geo-diverse visual concepts. 

Although the visual characteristics of the geo-diverse visual concepts in GIVL's pre-training corpus may be poles apart, they could be clustered in similar categories. For example, in Figure~\ref{intro-fig-1}, the visual characteristics of traditional Western and non-Western festivals are different, but these scenarios all belong to the same category \textit{festival}. 
Learning to identify category similarity can connect diverse visual concepts under similar categories and generalize to understanding more relevant concepts across regions more simply.
On the other hand, each of the visual concepts in similar categories associates with unique knowledge. Therefore, it is also crucial for GIVL to precisely distinguish if input knowledge aligns with the input image.

To this end, we construct the three types of input contents and formulate \ac{ikm} as a 3-way classification task to enforce GIVL to identify the input type:
\begin{itemize}
    \item Type 1: $\mathbf{k}$ matches input image $\mathbf{I}$;
    \item Type 2: $\mathbf{k}$ mismatches input image $\mathbf{I}$ and the visual concept described by $\mathbf{k}$ does \textbf{NOT} fall into a similar category of the visual concept $p_v$ in $\mathbf{I}$;
    \item Type 3: $\mathbf{k}$ mismatches input image $\mathbf{I}$ but the visual concept described by $\mathbf{k}$ falls into a similar category of the visual concept $p_v$ in $\mathbf{I}$. 
\end{itemize}
 

To select knowledge $\mathbf{k}$ for Type 3 input in \ac{ikm}, we need to conduct two steps (i) extracting the name of visual concept $p_v$ of input image $\mathbf{I}$ from its caption (for GQA data, see supplementary) and (ii) looking for visual concepts under similar categories. More details of extracting $p_v$ from image caption and its category information will be introduced in Section~\ref{sec:category}. After the visual concept name $p_v$ is extracted from the caption, to find a visual concept which falls in the most relevant categories to $p_v$, we randomly sample 200 visual concepts as candidates. Then we select the candidate concept that has the most semantically similar category with $p_v$'s category. Specifically, the sampled candidates are ranked by cosine similarity between text embeddings\footnote{We utilize FastText~\cite{bojanowski-etal-2017-enriching} embeddings pre-trained on Wikipedia. Phrase embeddings are mean pooled embeddings of the words in the phrases.} of their category names and the embedding of $p_v$'s category name. The process of selecting the most relevant visual concept $p^*$ is illustrated in Eq.~\eqref{eq:select},
\begin{equation}
   p^* = \argmaxA_{p_i} \mathrm{\mathbf{CosineSim}}(z_{p_i}, z_{p_v}),
\label{eq:select}
\end{equation}
where $z_{p_i}$ is the embedding of $i$-th sampled visual concept $p_i$'s category, $z_{p_v}$ is the embedding of $p_v$'s category, $\mathrm{\mathbf{CosineSim}}$ denotes the function quantifying cosine similarity between two embeddings. The corresponding knowledge of $p^*$ can be regarded as $\mathbf{k}$ in Type 3 input. 

For preparing $\mathbf{k}$ in Type 2 input content, we first set up a threshold for the cosine similarity between the embeddings of category names ($\tau=0.3$) to filter out the visual concepts relevant with $p_v$. Then we randomly pick one of the retained visual concepts. The selected visual concept indicates the one that has irrelevant category information with $p_v$. Its corresponding knowledge can be used as $\mathbf{k}$ in Type 2 input. 

\ac{ikm} loss is a cross-entropy loss with respect to the probability of predicting the type of relationship between the input image and knowledge upon $\mathrm{[CLS]}$ representation,
\begin{equation}
    \mathcal{L}_{IKM} = - \frac{1}{|\mathcal{D}|}\sum_{i=1}^{|\mathcal{D}|} \log p(y^k_{i} | \mathbf{c},\mathbf{k},\mathbf{t},\mathbf{v}),
    \label{eq:ikm}
\end{equation}
where $\mathcal{D}$ indicates the entire pre-training corpus\footnote{The proportion of Type 1, 2 and 3 input for IKM in $\mathcal{D}$ is $2:1:1$.} and $y^k_{i}$ is the label for the input type in IKM.

\subsubsection{Image Edit Checking (IEC)}
\label{sec:iec}
To better differentiate visually similar but irrelevant concepts, we propose another pre-training objective \acf{iec}. In geo-diverse setting, it is highly likely that visual concepts share similar visual characteristics but fall into completely different categories. For example, in Figure~\ref{intro-fig-2}, \textit{Chinese paper cuttings} are red and circular, which aligns with the visual characteristics of \textit{red frisbee}. \ac{iec} is designed to identify whether a specific visual concept $p_v$ in input image $\mathbf{I}$ is replaced by another visually similar one in an irrelevant category. 

We consider two types of input contents for \ac{iec}:
\begin{itemize}
    \item Type 1: Input image $\mathbf{I}$ remains the same;
    \item Type 2: The visual embedding of the visual concept $p_v$ in input image $\mathbf{I}$ is replaced with the embedding of another concept that is visually similar but falls into an irrelevant category with $p_v$.
\end{itemize}
In Figure~\ref{fig:model}, since the visual concept \textit{Chinese paper cuttings} is replaced with \textit{red frisbee}, the input type is Type 2\footnote{The proportion of Type 1 and 2 input for IEC in $\mathcal{D}$ is $1:1$.}.

\begin{figure}[t]
    \centering
    \includegraphics[width=0.83\linewidth, trim=0 10 0 0, clip]{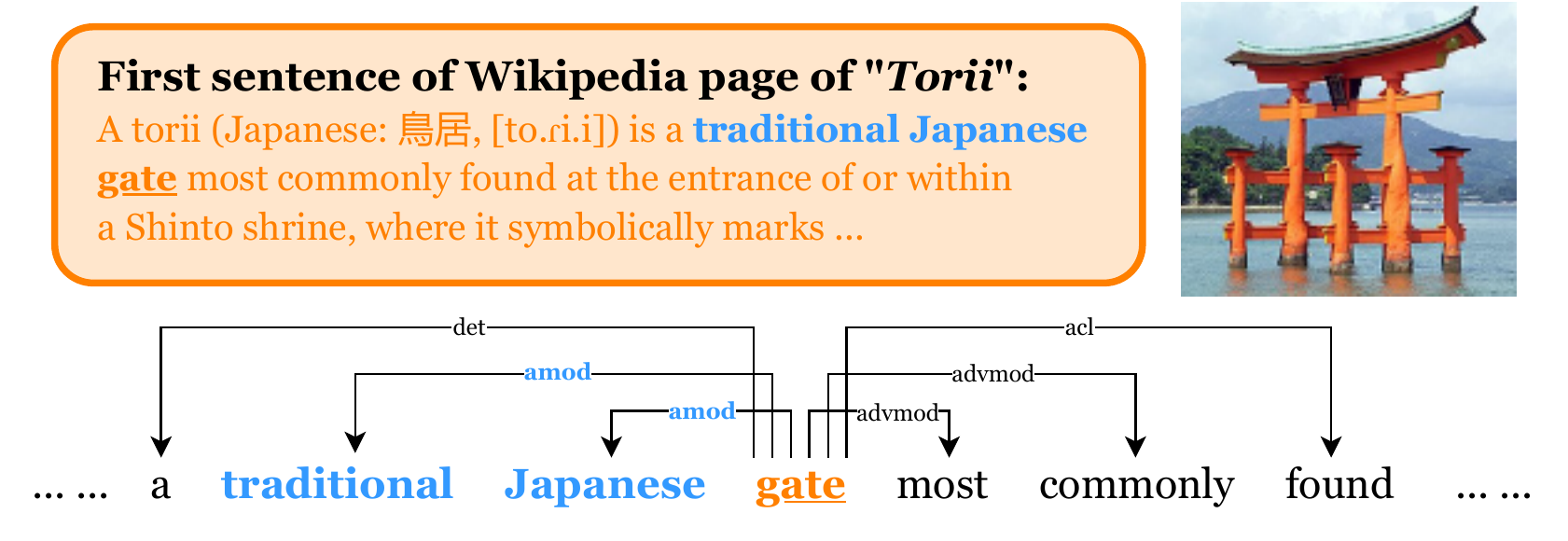}
    \caption{Steps to mine the category information of visual concepts. 
    The composition of the head noun (``\textit{gate}'', root of parse tree) and its modifiers (``\textit{traditional Japanese}'', words with ``amod'' relation with ``\textit{gate}'') can be treated as the category of \textit{torii} (``\textit{traditional Japanese gate}'').}
    \label{fig:category}
    \vspace{-8pt}
\end{figure}

To prepare input contents of Type 2 data, we need to accomplish two steps (i) seeking the corresponding detected objects of the visual concept $p_v$ in input image $\mathbf{I}$ from its caption (ii) looking for visually similar concepts for replacement. The pipeline of locating visual concept $p_v$ is introduced in Section~\ref{sec:locate}. After the visual concept $p_v$ is located, to select the proper visual concept for replacement in Type 2 input, we randomly sample 20 images, and then collect the visual embeddings and tag names of all the detected objects in the sampled images as candidates. The visual concept for replacement is selected according to two criteria: (i) its category is dissimilar\footnote{We use the cosine similarity between embeddings of the candidate visual concept's category and $p_v$'s category. Any candidate concepts with a similarity lower than 0.3 are treated as dissimilar ones.} with the category information of concept $p_v$ and (ii) its visual embedding is closest to $p_v$'s visual embedding. We select irrelevant visual concepts with $p_v$ to guarantee that the replacement is unreasonable given the image context. 

\ac{iec} loss is a binary cross-entropy loss with respect to the probability of predicting whether the input image is modified upon the $\mathrm{[CLS]}$ representation,
\begin{equation}
  \mathcal{L}_{IEC} = - \frac{1}{|\mathcal{D}|}\sum_{i=1}^{|\mathcal{D}|} \log p(y^v_{i} | \mathbf{c},\mathbf{k},\mathbf{t},\mathbf{v}),
  \label{eq:iec}
\end{equation}
where $y^v_{i}$ is the label for input type in IEC.
The final loss $L$ is the sum of all losses mentioned above:
\begin{equation}
  \mathcal{L} = \mathcal{L}_{MLM} + \mathcal{L}_{ITM} + \mathcal{L}_{IKM} + \mathcal{L}_{IEC}.
  \label{eq:loss}
\end{equation}

\subsection{Acquiring Categories of Visual Concepts}
\label{sec:category}

Acquiring the categories of visual concepts is a prerequisite step to construct GIVL inputs for IKM and IEC. We first need to extract the visual concept name $p_v$ in input image $\mathbf{I}$ from its image caption. We achieve this by parsing the caption with \cite{qi-etal-2020-stanza}. $p_v$ is the composition of the head noun and its modifiers in the parse tree. For example, given a caption ``\textit{Chinese paper cuttings in a shop}'', $p_v$ is \textit{Chinese paper cuttings}, which is composed of the head noun ``\textit{cuttings}'' and its modifiers ``\textit{Chinese paper}'' in its parse tree. 


To acquire $p_v$'s category, we then search for Wikipedia with keyword $p_v$. If $p_v$ is an entry of Wikipedia, we find that the category information can be usually found in the first sentence of Wikipedia introduction paragraph. As shown in Figure~\ref{fig:category}, the category of \textit{torii} (i.e., \textit{traditional Japanese gate}) is present in the first sentence ``\textit{A torii ... is a traditional Japanese gate most commonly ...}'' Then we notice that the category name is the phrase consisting of the head noun and its modifiers in the first sentence. In this example, the head noun of the first sentence is ``\textit{gate}'' and its modifier words are ``\textit{traditional}'' and ``\textit{Japanese}''. The final concatenation, ``\textit{traditional Japanese gate}'', is the category of \textit{torii}. Though the category information mined with these simple heuristics is imperfect, the extraction method is easy to implement and efficient in acquiring categories of large quantities of visual concepts.

\begin{figure}[t]
    \centering
    \includegraphics[width=0.85\linewidth]{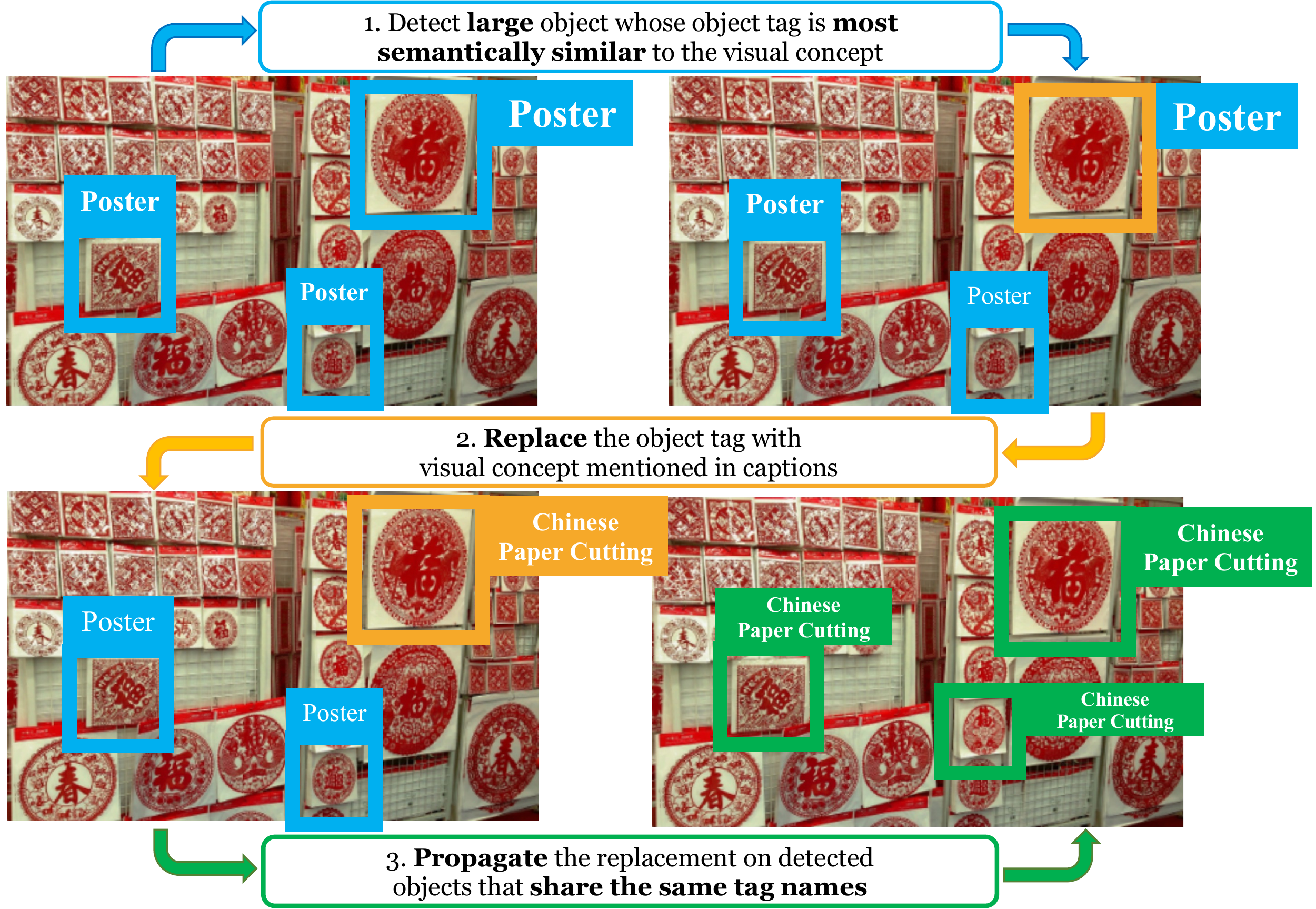}
    \caption{Steps to locate novel visual concepts in input images.}
    \label{fig:locate}
    \vspace{-8pt}
\end{figure}

\subsection{Locating Visual Concepts in Images}
\label{sec:locate}

With a limited amount of object class labels, it is difficult for current object detectors to detect a geo-diverse visual concept $p_v$. Therefore, we introduce a simple approach to efficiently locate the corresponding object given a visual concept $p_v$. We find that a visual concept $p_v$ is commonly (i) classified as a tag name that has similar semantics with $p_v$'s category, and (ii) its image patch occupies a large portion of the image. To this end, we design heuristics to locate novel visual concepts according to our empirical findings. First, only the top-10 large detected objects from each image will be considered. Second, we calculate the similarity between their object tags and $p_v$'s category. The one with the highest similarity score will be treated as the object corresponding to $p_v$. We take Figure~\ref{fig:locate} as an example. The visual concept $p_v$ to be located is \textit{Chinese paper cuttings}. Suppose that one of the \textit{Chinese paper cuttings} (the object in top right corner) is among top-10 large detected objects. Besides, its original detected object tag is \textit{poster}, which is the most semantically similar to \textit{Chinese paper cuttings}'s category. Hence, we can replace its original object tag with \textit{Chinese paper cuttings} as it is the corresponding object we are looking for.



The method above only locates one visual concept per image. However, it is possible that one image may contain multiple identical visual concepts. For example, in Figure~\ref{fig:locate}, there are a couple of \textit{Chinese paper cuttings}. To solve this problem, we simply propagate the visual concept name of \textit{Chinese paper cuttings} to other objects that share the same original detection labels.

%% file: sections/experiments.tex
\vspace{-3pt}
\section{Experiments} \label{sec:experiment}

We conduct two sets of experiments to evaluate GIVL. We first evaluate GIVL on multiple geo-diverse \ac{vl} tasks including zero-shot image classification, \ac{vl} reasoning and image-text retrieval. It helps us to verify the effectiveness of GIVL under geo-diverse settings. On the other hand, experiments on common \ac{vl} tasks are conducted to prove the generalizability of GIVL's pre-training method. 

\subsection{Baselines for Ablation Study}
\label{sec:baselines}
Five baselines are described below. For fair comparison, pre-training corpus, number of pre-training steps, and hyper-parameters are all identical to GIVL\footnote{Details of experimental setups are described in Appendix~\ref{sec:finetuning}.}. Since \ac{vl} pre-training is extremely time consuming, all baselines are pre-trained with 500K steps in ablation study.

\vspace{-12pt}
\paragraph{GIVL w/o $\mathcal{L}_{IKM}$ \& GIVL w/o $\mathcal{L}_{IEC}$.} GIVL w/o $\mathcal{L}_{IKM}$ and GIVL w/o $\mathcal{L}_{IEC}$ is the model pre-trained without Image-Knowledge Matching (IKM) objective and Image Edit Checking (IEC) objective, respectively. We demonstrate the effectiveness of our proposed pre-training objectives with these two baselines.

\vspace{-12pt}
\paragraph{VinVL$^\mathbf{*}$.} VinVL$^\mathbf{*}$ is pre-trained only with \ac{mlm} and \ac{itm} objectives as VinVL\cite{zhang2021vinvl}. It also shares the same pre-training corpus with GIVL. The only difference between GIVL and VinVL$^\mathbf{*}$ is objectives. GIVL is pre-trained with \acf{ikm} and \acf{iec} but VinVL$^\mathbf{*}$ is not. 
Comparing GIVL and VinVL$^\mathbf{*}$ can manifest the improvement by introducing \ac{ikm} and \ac{iec} objectives on geo-diverse \ac{vl} tasks. The comparison is also fair for the pre-training methods of GIVL and VinVL on common \ac{vl} tasks.

\vspace{-12pt}
\paragraph{GIVL w/ CLIP.} Some recent \ac{vlp} utilize CLIP~\cite{radford2019language} as the vision encoder. We replace object-level visual encoder in GIVL with CLIP to check if it can further improve performance. CLIP provides grid-level visual representation instead of object-level's. Therefore, \ac{iec} objective is removed because it involves object-level replacements.

\vspace{-12pt}
\paragraph{GIVL-B.} The only difference between GIVL and GIVL-B is that the \ac{ikm} objective of GIVL-B is a binary classification objective instead of 3-way classification. For \ac{ikm}, it requires GIVL-B to identify whether the input knowledge matches the image contents. GIVL-B doesn't need to judge whether the input knowledge describes a visual concept that shares similar category with the concept in input image. The comparison between GIVL and GIVL-B is able to demonstrate the effect of incorporating category information for learning the knowledge of geo-diverse visual concepts.

\begin{table}[t]
    \centering
    \scalebox{0.73}{
        \begin{tabular}{l c c c}
        \toprule
        \textbf{Model}                              & \#Param       & Acc.             & Western/non-Western    \\
        \hline
        \textbf{Prior VLPs}                              &               &                 &                   \\
        \hline
        VinVL$^\mathbf{*}$                          & 112M          & 1.21            & 1.77/1.01         \\ 
        VinVL~\cite{zhang2021vinvl}                 & 112M          & 1.29            & 1.25/1.30         \\
        \hline
        \textbf{Ours}                               &               &                 &                   \\
        \hline
        GIVL w/o $\mathcal{L}_{IKM}$                & 112M          & 21.37           & 25.31/20.37         \\
        GIVL w/o $\mathcal{L}_{IEC}$                & 112M          & 12.96           & 12.71/13.02         \\
        GIVL w/ CLIP                                & 199M          & 18.04           & 22.89/16.82          \\
        GIVL-B                                      & 112M          & 20.35           & 23.93/19.45          \\
        GIVL                                        & 112M          & \textbf{27.25}  &\textbf{31.65}/\textbf{26.15} \\
        \bottomrule
    \end{tabular}
    }
    \caption{Results on geo-diverse zero-shot image classification on Dollar Street dataset. We also show the respective performance on Western and non-Western images.}
    \label{tbl:dollar_street}
    \vspace{-4pt}
\end{table}

\begin{table}[t]
    \centering
    \scalebox{0.73}{
        \begin{tabular}{l c c c c}
            \toprule
            \textbf{Model}                       & Data/Steps         & \#Param       & Acc.       & $\Delta$   \\
            \hline
            \textbf{Prior VLPs}                       &                    &               &            &                   \\
            \hline
            ViLBERT~\cite{lu2019vilbert}         & 3.3M/-             & 274M          & 66.53      & 10.87             \\
            VinVL~\cite{zhang2021vinvl}          & 5.65M/2M           & 112M          & 72.48      & 8.55              \\
            VinVL$^*$                            & 3.17M/500K         & 112M          & 69.66      & 8.27                 \\
            X-VLM$\dagger$~\cite{zeng2021multi}           & 16M/-              & 216M          & 73.02      & 11.39             \\
            ALBEF$\dagger$~\cite{li2021align}             & 14M/-              & 210M          & 73.17      & 9.37              \\
            METER$\dagger$~\cite{dou2022empirical}        & 4M/-               & 352M          & 73.47      & 8.86              \\
            \hline
            \textbf{Ours}                        &                    &               &               \\
            \hline
            GIVL w/o $\mathcal{L}_{IKM}$         & 3.17M/500K         & 112M             & 72.11           & -         \\
            GIVL w/o $\mathcal{L}_{IEC}$         & 3.17M/500K         & 112M             & 68.58           & -         \\
            GIVL w/ CLIP                         & 3.17M/500K         & 199M             & 71.78           & -         \\
            GIVL-B                               & 3.17M/500K         & 112M             & 70.26           & -         \\
            GIVL                                 & 3.17M/500K         & 112M             & 72.50           & \textbf{6.56}      \\
            \hline
            GIVL                                 & 3.17M/900K         & 112M             & \textbf{72.70}  & 7.17      \\
            \bottomrule
        \end{tabular}
    }
    \caption{Results on MaRVL testing set. We also show the performance discrepancy $\Delta$ between NLVR2 and MaRVL. $\dagger$ denotes the results reported in \cite{zhou2022vlue}.}
    \label{tbl:marvl}
\end{table}

\begin{table}[t]
    \centering
    \scalebox{0.73}{
        \begin{tabular}{l c c c c}
            \toprule
            \textbf{Model}                              & \#Param       & Acc.             & Non-West            & $\Delta$            \\
            \hline
            \textbf{Prior VLPs}                              &               &                 &                     &       \\
            \hline
            VisualBERT~\cite{lu2019vilbert}             & 135M          & 53.95           & -                   & 10.42     \\
            ViLBERT~\cite{lu2019vilbert}                & 274M          & 59.99           & -                   & 7.28      \\
            VinVL$^\mathbf{*}$                          & 112M          & 69.07           & 66.45               & 8.46      \\ 
            VinVL~\cite{zhang2021vinvl}                 & 112M          & 70.20           & 66.78               & 11.04     \\
            \hline
            \textbf{Ours}                               &               &                 &                     &           \\
            \hline
            GIVL w/o $\mathcal{L}_{IKM}$                & 112M          & 69.56           & 65.96               & -     \\
            GIVL w/o $\mathcal{L}_{IEC}$                & 112M          & 69.89           & 66.92               & -      \\
            GIVL w/ CLIP                                & 199M          & 70.43  & 67.25               & 10.20      \\
            GIVL-B                                      & 112M          & 69.56           & 65.96               & -     \\
            GIVL                                        & 112M          & 70.32           & 68.41       & 6.14      \\
            \hline
            GIVL (1M)                                   & 112M          & \textbf{72.01}  &\textbf{70.4}        & \textbf{4.97}      \\
            \bottomrule
        \end{tabular}
    }
    \caption{Results on GD-VCR. We also show the results on all the non-Western images in GD-VCR and discrepancy $\Delta$ between Western and non-Western images.}
    \label{tbl:gdvcr}
    \vspace{-6pt}
\end{table}

\vspace{-3pt}
\subsection{Results on Geo-Diverse Benchmarks} \label{sec:exp_geo}
\paragraph{Geo-Diverse Zero-Shot Image Classification.}
Geo-diverse zero-shot image classification is a downstream geo-diverse \ac{vl} task that directly evaluates the effectiveness of the pre-training methods. 
We evaluate models on Dollar Street dataset\footnote{Images in Dollar Street are labeled with country information. The proxy to categorize Western and non-Western countries is based on \cite{huntington2000clash}.}. It is labeled with 127 classes, each of which contains images around the world. 
For classification on one image, we compose 127 inputs, each of which is the concatenation of one class name, the class's corresponding knowledge\footnote{The knowledge of each class sources from Wikipedia and Wordhoard.}, tag names and visual embeddings of the detected objects. We compare the probability of predicting that each class name matches the input image via \ac{itm} objective for all the 127 classes. The class with the highest probability is treated as the final classification result.

As shown in Table \ref{tbl:dollar_street}, GIVL outperforms both VinVL and VinVL$^\mathbf{*}$ by a significant margin around 26\%. GIVL achieves 6\%-20\% improvement in ablation studies, demonstrating the effectiveness of the proposed \ac{ikm} and \ac{iec} objectives. We also find that GIVL outperforms GIVL w/ CLIP which involves a strong vision encoder. It further demonstrates that object-level visual representations and object-level pre-training objective \ac{iec} are effective for learning geo-diverse visual concepts.

\vspace{-9pt}
\paragraph{Multicultural Visual Reasoning (MaRVL).}
Following NLVR2\cite{suhr-etal-2019-corpus}, MaRVL\cite{liu-etal-2021-visually} is a \ac{vl} task that requires models to identify whether a sentence correctly describes the contents of two input images. MaRVL images involve diverse visual concepts in non-Western regions. Since MaRVL\footnote{We use the translated English version of MaRVL dataset in \cite{zhou2022vlue}.} is merely a testing set, following \cite{liu-etal-2021-visually}, we fine-tune models on NLVR2 training set and then select the best checkpoint on the dev set of NLVR2 to evaluate on MaRVL.

From Table \ref{tbl:marvl}, we observe that GIVL outperforms the ablated baselines pre-trained without our proposed objectives IKM and IEC, respectively. Also, similar to the observations on Dollar Street dataset, compared with VinVL$^*$ pre-trained with the same corpus as GIVL, GIVL achieves higher performance. It further demonstrates that the pre-training objectives of GIVL can help VLPs learn geo-diverse visual concepts better than VinVL.

We also compare GIVL with \ac{vlp} (i) $3\times$ larger model (METER) and (ii) pre-trained with $2-5\times$ larger corpus (VinVL, X-VLM and ALBEF). GIVL achieves competitive performance with much less data and smaller model size. Additionally, we attach importance to the comparison of GIVL between NLVR2 and MaRVL. \cite{liu-etal-2021-visually} demonstrate that the visual concepts in NLVR2 dataset are Western-centric. A smaller performance gap between NLVR2 and MaRVL means less bias against non-Western regions. We observe that GIVL can achieve more balanced performance on both datasets, while other \ac{vlp} including METER, X-VLM and ALBEF have larger performance discrepancy. 

\vspace{-9pt}
\paragraph{Geo-Diverse Visual Commonsense Reasoning (GD-VCR).} 
GD-VCR is a testing set to evaluate multi-modal models’ ability to understand geo-diverse commonsense knowledge in images. It is a multiple-choice QA task which requires geo-diverse commonsense reasoning. We fine-tune models on VCR~\cite{zellers2019recognition} training set and select the best checkpoint on VCR's dev set to evaluate on GD-VCR.


As shown in Table \ref{tbl:gdvcr}, GIVL outperforms all prior similar-size \ac{vlp} models trained with similar number of images. GIVL also outperforms all ablated baselines except for GIVL w/ CLIP, which uses a much stronger visual encoder and only achieves 0.1\% subtle improvements. Besides, we highlight the performance gap between Western and non-Western data in GD-VCR. GIVL has significantly smaller gap than any of the ablated baselines. While GIVL w/ CLIP has a marginal improvement over GIVL, the performance gap of GIVL is 4.06\% smaller than GIVL w/ CLIP.

\begin{table}[t]
    \centering
    \scalebox{0.73}{
        \begin{tabular}{l c c c c}
            \toprule
            \textbf{Model}                       & Data         & \#Param       & I/R            & T/R      \\
            \hline
            \textbf{Prior VLPs}                       &                    &               &                &          \\
            \hline
            LXMERT~\cite{tan2019lxmert}          & -                & 240M          & 14.28          & 14.86    \\
            VisualBERT~\cite{li2019visualbert}   & -                & 135M          & 15.36          & 15.75    \\
            UNITER~\cite{chen2020uniter}         & -                & -             & 15.43          & 16.01    \\
            VL-BERT~\cite{su2019vl}              & 3.3M             & -             & 15.11          & 16.09    \\
            ViLBERT~\cite{lu2019vilbert}         & 3.3M             & 274M          & 15.40          & 16.93    \\
            VinVL~\cite{zhang2021vinvl}          & 5.65M         & 112M          & 27.78          & 28.65    \\ 
            VinVL$^*$                            & 3.17M           & 112M             & 25.44          & 25.50    \\
            \hline
            \textbf{Ours}                        &                    &               &                &          \\
            \hline
            GIVL w/o $\mathcal{L}_{IKM}$         & 3.17M         & 112M          & 26.21          & 26.97    \\
            GIVL w/o $\mathcal{L}_{IEC}$         & 3.17M         & 112M          & 28.08          & 28.18    \\
            GIVL w/ CLIP                         & 3.17M         & 199M          & 27.94          & 28.17          \\
            GIVL-B                               & 3.17M         & 112M          & 29.97          & 29.86          \\
            GIVL                                 & 3.17M         & 112M          & 28.00          & 28.79    \\
            \hline
            GIVL (1M)                                & 3.17M           & 112M          & \textbf{29.98} & \textbf{30.79}    \\
            \bottomrule
        \end{tabular}
    }
    \caption{Results on WIT image-text retrieval task. I/R and T/R denote image retrieval and text retrieval. The evaluation metric is Recall@1. 1M denotes the number of pre-training steps.}
    \label{tbl:wit}
\end{table}

\begin{figure}[t]
    \centering
    \includegraphics[width=0.93\linewidth, trim=0 0 0 0, clip]{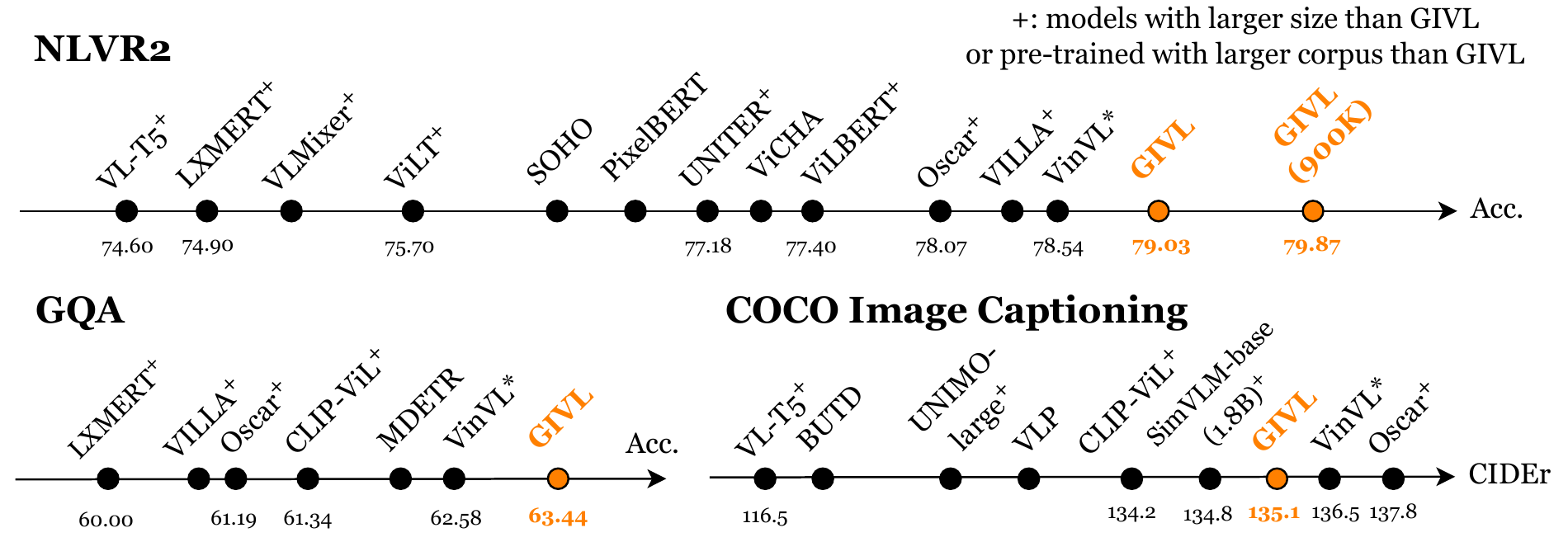}
    \caption{GIVL performance on common V\&L tasks. Complete results are shown in Appendix~\ref{sec:detail_results}.}
    \label{fig:common}
    \vspace{-6pt}
\end{figure}

\paragraph{Wikipedia Image-Text Retrieval (WIT).}
WIT image-text retrieval is a standard retrieval task on geo-diverse Wikipedia images\footnote{We use the translated English WIT retrieval data in \cite{bugliarello-etal-2022-iglue}.}. Table~\ref{tbl:wit} shows that GIVL achieves superior performance comparing to baselines except GIVL-B. Pre-trained with 1M steps, GIVL obtains SOTA performance on WIT image-text retrieval task.

\subsection{Results on Common V\&L Benchmarks}
Besides testing GIVL on geo-diverse \ac{vl} tasks, we benchmark GIVL on common \ac{vl} task to investigate whether the pre-training method of GIVL is competitive among existing \ac{vlp}. We don't expect GIVL to perform the best among SOTA \ac{vlp} on these \ac{vl} benchmarks, because they are annotated with Western-centric data and SOTAs are trained with much larger similar data as well.
We aim to answer two questions. Q1: \emph{Is GIVL able to obtain comparable performance with \ac{vlp} pre-trained with similar scale of data?} Q2: \emph{Can GIVL perform as strongly as SOTA \ac{vlp} pre-trained with the same corpus?}

\begin{figure}[t]
    \centering
    \includegraphics[width=0.95\linewidth]{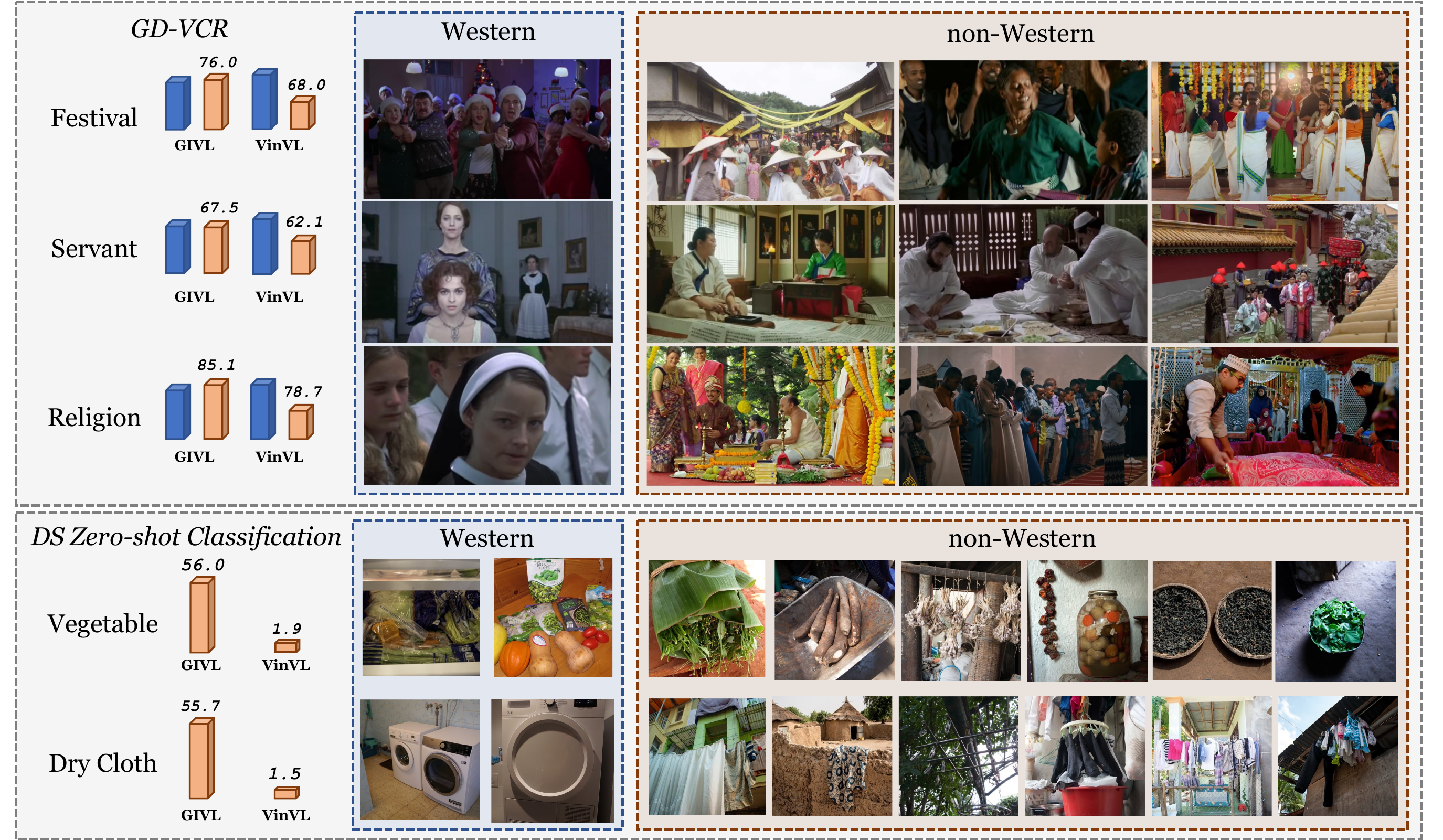}
    \caption{GIVL and VinVL's performance on non-Western and Western images related to geo-diverse categories.}
    \label{fig:showcase}
    \vspace{-6pt}
\end{figure}

To answer Q1, we evaluate GIVL on common \ac{vl} benchmarks including NLVR2, GQA and COCO captioning. For NLVR2, GIVL is able to beat 11 \ac{vlp} with much more parameters and pre-trained with more data. For GQA, GIVL performs better than most of the \ac{vlp}. 
For COCO image captioning, it can even obtain close performance with SimVLM-base, a VLP pre-trained with 1.8B images. Overall, even though GIVL is pre-trained with the corpus whose domain is not similar with common \ac{vl} benchmarks, it can still achieve competitive results. It demonstrates the effectiveness of GIVL pre-training method.

For Q2, we target on VinVL, a strong VLP that once swept leaderboards of multiple \ac{vl} tasks. For fair comparison, we reproduce the pre-training process of VinVL with GIVL pre-training corpus. As mentioned in Section~\ref{sec:baselines}, we denote the reproduced pre-training as VinVL$^*$. On above three \ac{vl} datasets, the performance difference between GIVL and VinVL$^*$ is subtle. We argue that GIVL could achieve equally good performance as VinVL on common \ac{vl} benchmarks if it was pre-trained with VinVL corpus. 

\subsection{Qualitative Study on Geo-Diverse Categories}
We showcase examples from GD-VCR and Dollar Street dataset to better demonstrate GIVL's advantages. As shown in Figure~\ref{fig:showcase}, non-Western \textit{festivals}, \textit{servants} and \textit{religions} are quite different from those in Western regions. We find that GIVL's performance gap on the images involving these categories is significantly smaller than VinVL on GD-VCR. Moreover, GIVL's performance on non-Western images is 5-8\% higher than VinVL. For Dollar Street dataset, while the overall performance of GIVL is around 30\%, GIVL can achieve above 55\% accuracy when recognizing \textit{vegetables} and \textit{drying clothes} which greatly vary across data from worldwide. GIVL even outperforms VinVL 50\% on those categories. GIVL's strong performance on these highly geo-diverse categories further demonstrates its effectiveness.


%% file: sections/conclusion.tex
\section{Conclusion} \label{conclusion}
We propose GIVL, a geographically inclusive vision-and-language pre-trained model. GIVL achieves strong and more balanced results on multiple geo-diverse \ac{vl} tasks. It can also produce competitive performance on common \ac{vl} tasks. By proposing GIVL, we call upon researchers to devise methods that can further improve geographical inclusivity of \ac{vlp} and popularize their applications for all.

%% file: sections/appendix.tex
\appendix

\section*{Appendix}

\section{Pre-Training and Fine-Tuning\\Details of Downstream V\&L Tasks}
\label{sec:finetuning}
\subsection{Pre-Training Setups}
GIVL is initialized with pre-trained parameters of BERT-base model\cite{wolf-etal-2020-transformers}. It is pre-trained for at most 1M steps with a batch size of 720. The learning rate is $1e-4$ with linear decay. The maximum numbers of tokens in input texts and visual objects are 70 and 50, respectively. All the pre-training experiments for GIVL and ablated baselines are implemented with 8 A100 GPUs with 40GB GPU memory.

\subsection{Fine-Tuning Setups}
\paragraph{MaRVL and NLVR2.}
We fine-tune GIVL on NLVR2 for 20 epochs, with batch size 72 and learning rate $3e-5$. The maximum number of tokens in input texts and visual objects is 55. Because each sample has two input images, we include a maximum of 80 visual objects in the model input, with each image having a maximum of 40 input visual objects. As mentioned in Section~\ref{sec:exp_geo}, since MaRVL is a testing set following NLVR2's formulation, the fine-tuning results are based on the fine-tuning method discussed here.

\paragraph{GD-VCR.}
We fine-tune GIVL on original VCR dataset for 5 epochs, with batch size 128 and learning rate $3e-5$. For model input, we concatenate the question and four answer choices together, along with the visual embeddings of the input image. The maximum numbers of tokens and visual objects are 100 and 50, respectively.

\paragraph{WIT Image-Text Retrieval.}
We fine-tune GIVL on the WIT Image-Text retrieval training set for 20 epochs, with a batch size of 128 and a learning rate of $2e-5$. The maximum numbers of tokens in input texts and visual objects are 70 and 70, respectively. We use the translated English training and dev set provided in IGLUE\cite{bugliarello-etal-2022-iglue}.

\paragraph{COCO Captioning.}
We fine-tune GIVL on COCO captioning dataset for 60 epochs, with batch size 256 and learning rate $3e-5$ with Seq2Seq objective~\cite{sutskever2014sequence,cho-etal-2014-learning}. The maximum numbers of tokens in input texts and visual objects are 70 and 50, respectively. After that, we further optimize GIVL with the CIDEr metric for 75 epochs with a batch size of 64 and a learning rate of $2e - 6$. We use beam search with beam size 5~\cite{anderson2018bottom} to sample the generation results, and the maximum length of the generated captions is 20 words.

\paragraph{GQA.}
We fine-tune GIVL on GQA for 5 epochs, with batch size 128 and learning rate $5e-5$. The maximum numbers of tokens in input texts and visual objects are 165 and 45, respectively.

\section{Detailed Results on Common V\&L Tasks}
\label{sec:detail_results}
As mentioned in Section \ref{sec:experiment}, we also conduct experiments on common \acf{vl} tasks. We show detailed experimental results in Table \ref{tbl:gqa}, \ref{tbl:nlvr2} and \ref{tbl:coco_caption} for GQA, NLVR2 and COCO captioning, respectively.

In Table \ref{tbl:gqa}, we show that GIVL outperforms many prior \acf{vlp} on GQA. We emphasize that GIVL is trained with significantly less data than most of the prior VLPs, while GIVL also uses fewer parameters compared to these \ac{vlp}. For fair comparison, VinVL$^*$ uses the same pre-training data as GIVL.
\begin{table}[h!]
    \centering
    \scalebox{1}{
        \begin{tabular}{l c c c}
            \toprule
            \textbf{Model}                       & \#Param   & Data    & Acc.           \\
            \hline
            \textbf{Prior VLPs}                       &       &        &               \\
            \hline
            LXMERT~\cite{tan2019lxmert}          & 240M   & -       & 60.00         \\
            Oscar~\cite{li2020oscar}             & -      & 4.1M       & 61.19         \\
            CLIP-ViL~\cite{shen2021much}                             & 178M    & -             & 61.34         \\
            MDETR~\cite{kamath2021mdetr}                                & -     & -        & 62.48         \\ 
            VinVL$^*$~\cite{zhang2021vinvl}      & 112M      & 3.17M    & 62.58         \\
            \hline
            \textbf{Ours}                        &         &      &               \\
            \hline
            GIVL                                 & 112M     & 3.17M     & \textbf{63.44}\\
            \bottomrule
        \end{tabular}
    }
    \caption{Results on GQA test-dev set.}
    \label{tbl:gqa}
\end{table}

We also evaluate the proposed GIVL on the NLVR2 dataset. Similar to results in GQA, according to Table \ref{tbl:nlvr2}, GIVL also outperforms all the listed prior \ac{vlp} with much less pre-training data and smaller model size.

\begin{table}[h]
    \centering
    \scalebox{1}{
        \begin{tabular}{l c c c}
            \toprule
            \textbf{Model}                       & \#Param      & Data & Acc.           \\
            \hline
            \textbf{Prior VLPs}                       &         &      &               \\
            \hline
            VL-T5~\cite{cho2021unifying}         & 224M         & - & 74.60         \\
            LXMERT~\cite{tan2019lxmert}          & 240M         & - & 74.90         \\
            VLMixer~\cite{wang2022vlmixer}       & -            & 4M & 75.28         \\
            ViLT~\cite{kim2021vilt}              & 87M          & 4M & 75.70         \\
            PixelBERT~\cite{huang2020pixel}      & 114M         & - & 76.73         \\ 
            SOHO~\cite{huang2021seeing}          & -            & - & 76.37         \\
            UNITER~\cite{chen2020uniter}         & 300M         & 4M & 77.18         \\
            ViCHA~\cite{shukor2022efficient}     & -            & - & 77.27         \\
            ViLBERT~\cite{lu2019vilbert}         & 274M         & 3.3M & 77.40         \\
            Oscar~\cite{li2020oscar}             & -            & 4.1M & 78.07         \\
            VILLA~\cite{gan2020large}            & -            & 4M & 78.39         \\
            VinVL$^*$~\cite{zhang2021vinvl}      & 112M         & 3.17M & 78.54         \\
            \hline
            \textbf{Ours}                        &          &     &               \\
            \hline
            GIVL                                 & 112M       & 3.17M   & \textbf{79.03}\\
            GIVL (900K)                                & 112M       & 3.17M   & \textbf{79.87}\\
            \bottomrule
        \end{tabular}
    }
    \caption{Results on NLVR2 test-dev set.}
    \label{tbl:nlvr2}
\end{table}

\begin{table*}[t]
    \centering
    \scalebox{1}{
        \begin{tabular}{l c c c c c c}
            \toprule
            \textbf{Model}                       & \#Param       & Data   & BLEU@4   & CIDEr    & METEOR    & SPICE  \\
            \hline
            \textbf{Prior VLPs}                       &               &               \\
            \hline
            VL-T5~\cite{cho2021unifying}         & 224M          & -        & - & 116.5    & -         & -      \\
            BUTD~\cite{anderson2018bottom}                                 & -      & -       & 36.3     & 120.1    & 27.7      & 21.4   \\
            VLP~\cite{zhou2020unified}                                  & -     &   -      & 39.5     & 129.8    & 29.3      & 22.4   \\
            Unimo-Large~\cite{li2020unimo}                          & 300M     & -        & 39.6     & 127.7    & -         & -      \\
            Oscar~\cite{li2020oscar}                                & -       &  4.1M    & 40.5     & 137.6    & 29.7      & 22.8   \\
            CLIP-ViL~\cite{shen2021much}                             & 178M      &  -       & 40.2     & 134.2    & 29.7      & 23.8   \\
            SimVLM-base\cite{wang2021simvlm}	      & -    & 1.8B   & 39	& 134.8	& 32.9	& 24 \\
            VinVL$^*$~\cite{zhang2021vinvl}      & 112M     & 3.17M     & 39.6     & 136.5    & 30.4      & 24.4   \\
            \hline
            \textbf{Ours}                        &          &     &          &          &           &        \\
            \hline
            GIVL                                 & 112M      & 3.17M    & 39.6     & 135.1    & 30.3      & 24.3   \\
            \bottomrule
        \end{tabular}
    }
    \caption{Results on COCO captioning.}
    \label{tbl:coco_caption}
\end{table*}

\begin{figure*}[h]
    \centering
    \scalebox{0.91}{
    \includegraphics[width=\linewidth, trim=0 0 0 0, clip]{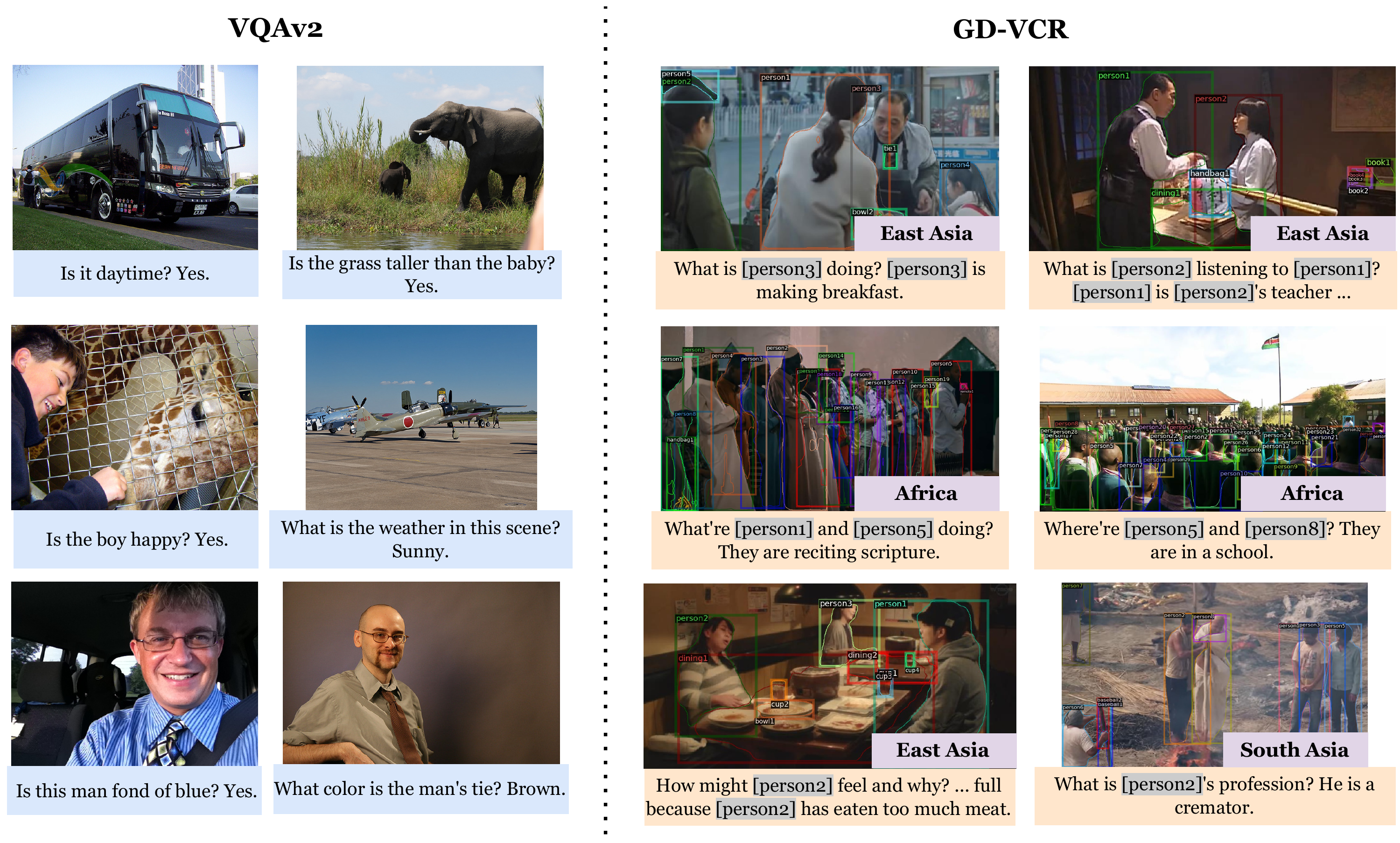}
    }
    \caption{Comparison between VQAv2 and GD-VCR's images and corresponding question-answer pairs.}
    \label{fig:vqa_gdvcr}
\end{figure*}

\begin{figure*}[h]
    \centering
    \scalebox{0.91}{
    \includegraphics[width=\linewidth, trim=0 0 0 0, clip]{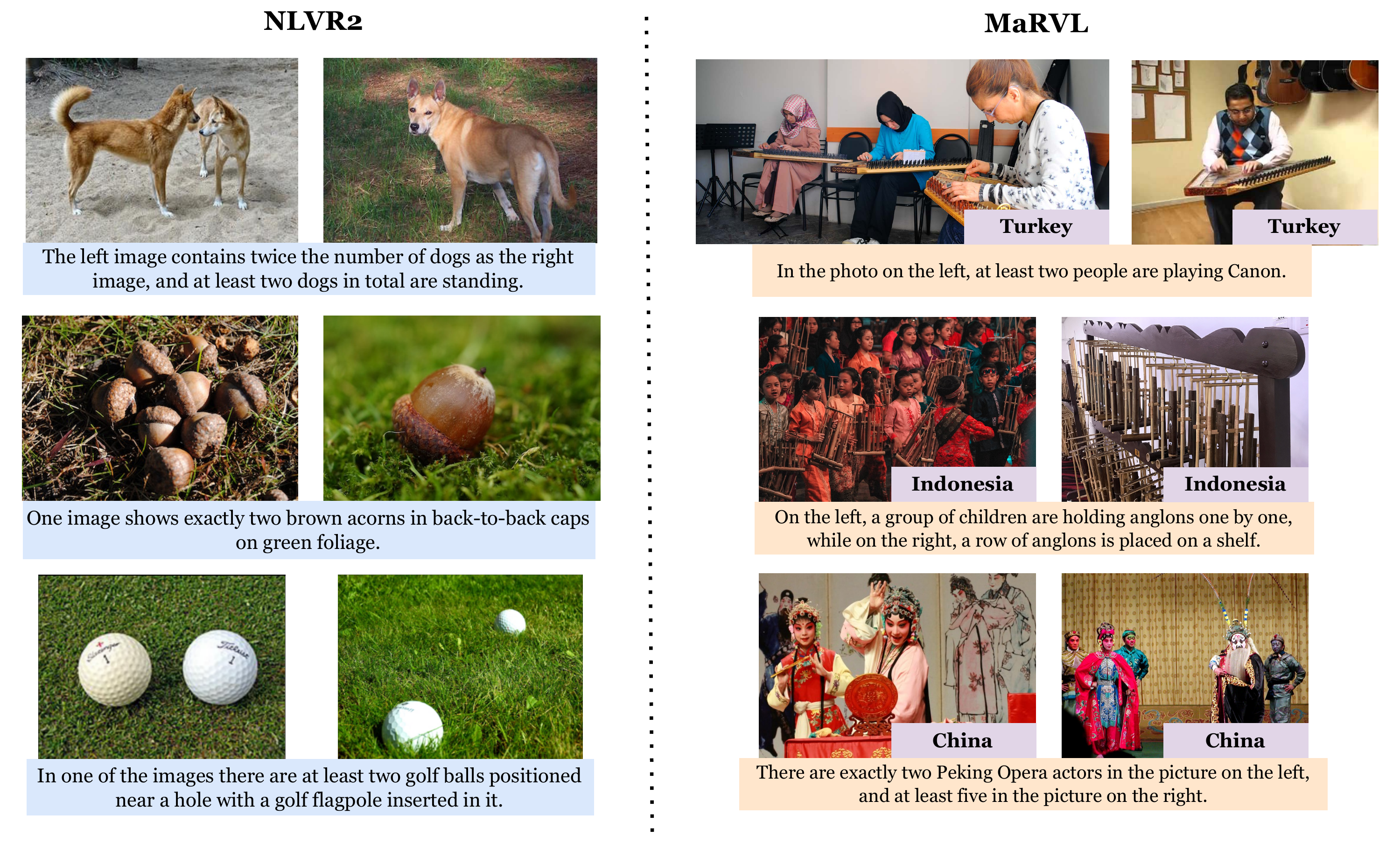}
    }
    \caption{Comparison between NLVR2 and MaRVL's images and claims.}
    \label{fig:nlvr2_marvl}
\end{figure*}

Image captioning is a classic task to evaluate the performance of \ac{vlp}. As illustrated in Table \ref{tbl:coco_caption}, GIVL shows comparable performance to prior \ac{vlp} in different evaluation metrics. Most of the prior image captioning \ac{vlp} use much more data than GIVL, for example, SimVLM-base. All three experiments above demonstrate the effectiveness and data efficiency of GIVL. 

\section{Qualitative Examples}
\subsection{Common v.s. Geo-Diverse V\&L Tasks}
Since geo-diverse \acf{vl} tasks are not widely studied in \acf{cv} community, it may not be intuitive enough for the audience to understand the differences between common \ac{vl} tasks and geo-diverse \ac{vl} tasks. In this section, we use some examples to illustrate it.

Before discussing the examples, we would like to introduce the setting of geo-diverse \ac{vl} tasks. First, geo-diverse \ac{vl} tasks, such as GD-VCR, only use images that are collected from different regions and cultures. It ensures that the visual concepts behind the images are highly relevant to the background regions and cultures. Second, these geo-diverse datasets require annotators from different regions and cultures to label the data, which further imposes the geo-diversity on them. Third and most importantly, questions or text descriptions in geo-diverse datasets focus more on the visual concepts from different regions and cultures and their corresponding knowledge.

Figure \ref{fig:vqa_gdvcr} shows some image-question pairs from both the VQA and GD-VCR datasets. The VQA dataset contains questions that ask for generic visual concepts, such as colors, weather, size, \etc. The visual information within the images of VQA dataset is sufficient to answer the questions. On the other hand, GD-VCR asks questions that require background knowledge from regions and cultures around the world. For example, the first example on the right-hand side describes a scenario where a person is making breakfast on a busy street. This is not a common occurrence in most Western countries, but it is very common in most East and South Asian regions. 

\subsection{Empirical Analysis of GIVL's Performances}
The comparison between VQA and GD-VCR also can indicate the reasons why GIVL has similar performances with other SOTAs on common \ac{vl} tasks but beats all baselines on geo-diverse tasks by a large margin. For common \ac{vl} tasks, although some images are collected around the world, they are not geo-diverse. Regardless of the geo-diverse factors in the image, the tasks only involve common visual concepts and their basic visual information. For instance, as shown in Figure \ref{fig:vqa_gdvcr}, the second image-question pair in the VQA examples only asks for the size information of elephants in the image. 
But the question doesn't ask for the implicit corresponding knowledge of tropical visual concepts. To this end, on common V\&L tasks, GIVL may not be able to outperform \ac{vlp} that are pre-trained with much greater \ac{vl} pre-training corpus mainly covering common visual concepts.

On the other hand, geo-diverse V\&L tasks such as GD-VCR and MaRVL, require models to complete the tasks with knowledge that is related to the background regions and cultures of the images. As shown in the right hand side of Figure \ref{fig:nlvr2_marvl}, the model needs to recognize geo-diverse visual concepts and leverage cultural knowledge beyond the image contents to make predictions. Since prior VLPs are not pre-trained to understand the underlying knowledge of geo-diverse visual concepts, GIVL can outperform the majority of SOTA \ac{vlp} on geo-diverse V\&L tasks. 

\subsection{Case Study of GD-VCR and MaRVL}
\begin{figure*}[h]
    \centering
    \scalebox{0.8}{
    \includegraphics[width=\linewidth, trim=0 0 0 0, clip]{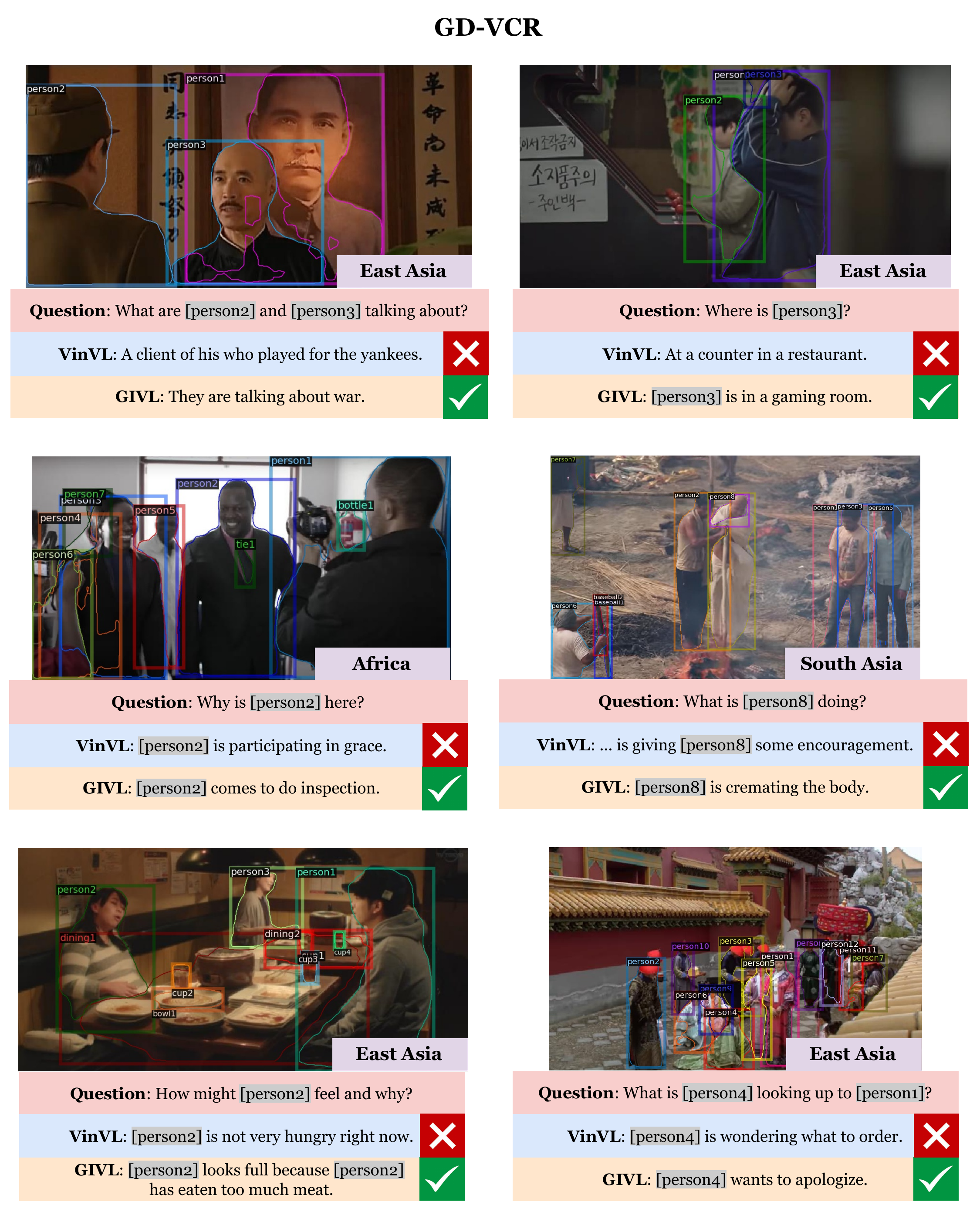}
    }
    \caption{Case study of GD-VCR.}
    \label{fig:gdvcr_case_study}
\end{figure*}

\begin{figure*}[t]
    \centering
    \scalebox{0.7}{
    \includegraphics[width=\linewidth, trim=0 0 0 0, clip]{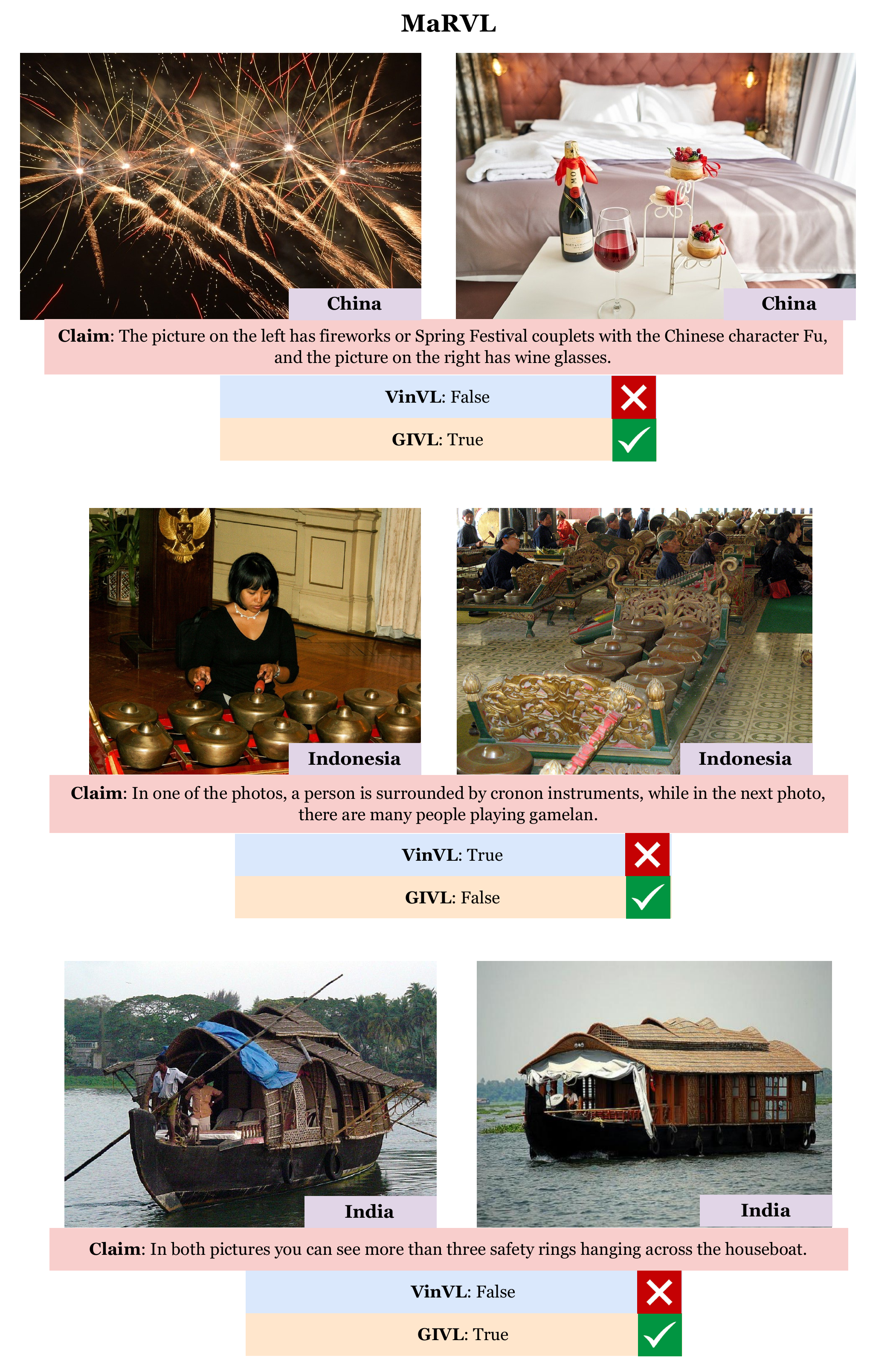}
    }
    \caption{Case study of MaRVL.}
    \label{fig:marvl_case_study}
\end{figure*}

We show some cases of GD-VCR and the predictions made by GIVL and VinVL in Figure \ref{fig:gdvcr_case_study}. VinVL is not able to solve some cases in GD-VCR while GIVL can reach the correct answers. In most shown cases, VinVL predictions do not make sense. These cases, such as the bottom-right example, are highly culturally related. People in that image wear ancient Chinese royal dress. The posture seems like they are lining up and half-squatting. In ancient China, it is a royal code for apology. 
More cases of MaRVL and the predictions of GIVL and VinVL are shown in Figure \ref{fig:marvl_case_study}.

